\documentclass[numbered]{trbunofficial}
\usepackage{graphicx}
\usepackage{amsmath}
\usepackage[hidelinks]{hyperref}

\AuthorHeaders{Prajapati, Singh, Hegde, and Chakraborty}
\title{Evaluation and Comparison of Visual Language Models for Transportation Engineering Problems}

\author{%
  \textbf{Sanjita Prajapati$^{1}$}\\
  Department of Civil Engineering,\\
  Indian Institute of Technology Kanpur, India\\
  \hfill\break
  \textbf{Tanu Singh$^{1}$}\\
  Department of Civil Engineering,\\
  Indian Institute of Technology Kanpur, India\\  
  \hfill\break%
  \textbf{Chinmay Hegde}\\
  Department of Computer Science and Engineering\\
  New York University, USA\\
  \hfill\break%
  \textbf{Pranamesh Chakraborty (Corresponding Author)}\\
  Department of Civil Engineering,\\
  Indian Institute of Technology Kanpur, India\\  
  pranames@iitk.ac.in\\
  \hfill\break%
  $^{1}$\textit{Equal contribution}
}

\begin{document}
\maketitle

\section{Abstract}
Recent developments in vision language models (VLM) have shown great potential for diverse applications related to image understanding. In this study, we have explored state-of-the-art VLM models for vision-based transportation engineering tasks such as image classification and object detection. The image classification task involves congestion detection and crack identification, whereas, for object detection, helmet violations were identified. We have applied open-source models such as CLIP, BLIP, OWL-ViT, Llava-Next, and closed-source GPT-4o to evaluate the performance of these state-of-the-art VLM models to harness the capabilities of language understanding for vision-based transportation tasks. These tasks were performed by applying zero-shot prompting to the VLM models, as zero-shot prompting involves performing tasks without any training on those tasks. It eliminates the need for annotated datasets or fine-tuning for specific tasks. Though these models gave comparative results with benchmark Convolutional Neural Networks (CNN) models in the image classification tasks, for object localization tasks, it still needs improvement. Therefore, this study provides a comprehensive evaluation of the state-of-the-art VLM models highlighting the advantages and limitations of the models, which can be taken as the baseline for future improvement and wide-scale implementation.

\hfill\break%
\noindent\textit{Keywords}: visual-language models, crack detection, congestion detection, helmet violation detection, image understanding
\newpage

\section{Introduction}

In recent years, there have been significant advancements in computer vision and language modeling for solving different tasks using deep learning. Inspired by the advancements in natural language processing using transformer based models, a new concept in computer vision called Vision Transformers (ViT) \cite{vlm}, was introduced in 2021 for image understanding. In the field of language modeling, many Large Language Models (LLMs) such as Llama and ChatGPT have shown excellent capability to solve a large variety of tasks. These models, which were initially designed for text inputs, now also support visual inputs, connecting vision to language and enabling zero-shot or few-shot learning. This development has the potential to unlock several applications that will be key to the current AI-based technological revolution. 

On the other hand, Convolutional Neural Networks (CNNs) have been extensively utilized for the past decade for vision-based processing, demonstrating efficient real-time performance. Their simpler architecture renders them favorable for real-time deployment. However, CNNs have limitations, such as the requirement for extensive datasets and the need for fine-tuning for almost every use case to achieve better results. This process is labor-intensive requiring manual annotations, highlighting the need for pre-trained models. This has created a necessity for foundational models that can be applied to various tasks without the need for extensive fine-tuning. 

In this study, our primary focus is to understand the capabilities and limitations of the state-of-the-art Vision Language Models (VLMs) in the field of vision-based transportation engineering tasks. The study involved the careful selection of three specific transportation engineering tasks, each of which presents distinct challenges and complexities. The first task focuses on detecting congestion on highways from surveillance cameras, which is a critical issue in transportation management. The second task involves the identification of cracks in pavement surfaces, an essential aspect of infrastructure maintenance. While both these tasks fall under the domain of image classification, we go further to understand capabilities of VLMs in object detection tasks.  The third task addresses the vital issue of detecting helmet violation, specifically determining whether motorbike riders were wearing helmets or not, which is crucial for enhancing safety on roadways. 

These tasks have been chosen due to the fact that they require fine-tuning of pre-trained CNN models. Notably, the classes relevant to these tasks are not included in the COCO dataset \cite{coco}, necessitating specialized attention and refinement. The performance of the chosen tasks has been rigorously evaluated, providing valuable insights into the effectiveness of zero-shot VLMs in addressing transportation engineering tasks.  In this research, both open-source and closed-source foundation models of VLM have been considered. Within the open-source category, the study delved into the performance of models such as CLIP \cite{clip}, BLIP \cite{li2022blip}, OWL-ViT \cite{OWL_vit}, and Llava-Next \cite{llava_next}. Furthermore, the study also included an analysis of the closed-source GPT-4o \cite{gpt4o} model, presenting a thorough evaluation of a range of foundational VLMs for transportation engineering tasks. 

The following section discusses the past studies on the application of VLM models in transportation. This is followed by the methodology used in both  image classification tasks, congestion and crack detection, and then addresses the methodology used in object detection. After that, the paper discusses the datasets used in this study and the results of our research. Finally, the conclusions and the future scope are highlighted. 

\section{Literature Review}


Large Language Models (LLMs) have revolutionized natural language processing (NLP), allowing machines to understand and generate human-like language with unprecedented success. The performance of LLMs in textual understanding and their versatility in different domains of language tasks has led to the exploration of multi-modal LLMs \cite{pan2024vlp}. Multimodal LLMs can process and generate information across various data types such as text, images, audio, and video. Vision Language Models (VLMs) blend computer vision and NLP capabilities. They are designed such that they can process and generate human-like text based on visual inputs, or the other way around \cite{vlm}. By bridging the gap between visual and textual data understanding, VLMs have various applications such as image captioning, visual question answering, textual descriptions, and even image generation.

Recently, pre-trained VLM with zero-shot prediction has attracted significant attention, where VLM is pre-trained on a large-scale image-text dataset. The pre-trained VLM with a rich textual and image understanding can then be directly applied to any visual task without fine-tuning. Zero-shot prediction implies that the model can interpret and generate descriptions or answer questions, based on textual instructions it has never seen before.

 \subsection{VLM in transportation}

 The VLMs and LLMs have lately demonstrated strong zero-shot capabilities and human-like reasoning capabilties. Recently, few studies have integrated the VLMs and LLMs for traffic-related tasks such as understanding traffic scenes, autonomous driving, and anomaly detection for enhancing interpretability, safety, and generalization capabilities. 

Some studies have attempted to leverage VLMs in autonomous driving for various purposes such as navigation, forecasting, interpreting vehicle action, and planning. DriveVLM employs VLM to interpret complex traffic scenarios for understanding and analyzing the scene to plan the actions for autonomous driving \cite{tian2024drivevlm}. Similarly, DriveGPT4, a multimodal LLM uses input multi-frame videos and textual queries to generate responses and predicts low-level control signals for vehicle action\cite{xu2023drivegpt4}. GPT in DriveGPT4 stands for Generative pre-trained transformer and The digit “4” represents multimodality. In Vision Language Planning (VLP), researchers also integrated language models with vision-based systems to enhance autonomous driving by improving their contextual understanding and generalization capabilities. It has two components, an agent-centric learning paradigm and a self-driving-car-centric learning paradigm that improves the local details in the BEV feature map and enhances the planning process respectively, by leveraging the knowledge encoded in the pre-trained language model \cite{pan2024vlp}. While these works focused on improving autonomous driving systems, in the domain of Visual Language Navigation, VLN system was developed to navigate action for intelligent vehicles leveraging LLM and VLM, by extracting landmark names from user’s language instructions, matching landmark names with environmental objects, and finally reasoning navigation actions for the intelligent vehicle \cite{hu2023VLN}. On the other hand, CityLLaVA, was developed to understand traffic scenarios in the city by fine-tuning VLMs by utilizing bounding-box guided view selection and prompt engineering modules \cite{wangcityllava}.

Apart from autonomous driving application and scene understanding, DriveCLIP \cite{hasan2024driveclip} framework explores the application of vision-language models, particularly the CLIP model, to identify distracted driving activities from naturalistic driving videos and images. This system offers zero-shot transfer, fine-tuning, and video-based models for driver’s state prediction. All these studies have only performed research on the homogeneous driving environment, consisting of major four-wheelers. The heterogeneous driving environments have not been explored by any researcher being complex containing different types of vehicles with varying speeds. 

Video Anomaly Detection is another field where few works have applied VLMs and LLMs for improved performances.VAD-LLaMA \cite{lv2024VAD} is such a framework where traffic anomaly is detected and localized in a long-range surveillance video. They have incorporated video-based large language models (VLLM) to make threshold-free detection and explain the reasons for the anomalies detected. They have also introduced a novel Long-Term Context module to alleviate the incapability of long-range context modeling in existing VLLMs. 

Except for anomaly detection work, the field explored by VLMs in transportation has used datasets primarily containing in-vehicle camera images or videos (except anomaly detection tasks). On the other hand, the surveillance camera based images/videos can also be analysed through VLMs for improved image/video understanding. Moreover, all these applications are focused on high-level tasks such as vehicle navigation, anomaly detection, etc. In addition to exploring VLMs in such high-level image and video understanding, there is also a need to understand and analyze the potential of VLMs in low-level image understanding tasks such as image classification and object detection. This will involve utilizing the vision and language modalities to significantly improve zero-shot or few-shot classification and detection tasks in the transportation domain.
We recognized these limitations in the application of VLMs in transportation engineering-related problems and applied different state-of-the-art vision-language models in basic image understanding tasks such as classification and object detection to understand the capabilities and limitations of the models.

\section{Methodology}
This study aims to leverage the capabilities of VLM for vision-based transportation engineering tasks which include; a) image classification and b) object detection. In this study, we tested the state-of-the-art VLMs for two image classification tasks a) congestion detection and b) crack detection. On the other hand, in the domain of object detection, we have evaluated the performance of VLM for detecting helmet violation cases i.e. motorbike riders wearing helmet or not.


These image classification and object detection tasks are selected to identify the potential of VLM in tasks that go beyond detecting regular traffic entities (such as cars, pedestrians, etc) and therefore can harness the capabilities of language understanding for vision-based transportation tasks. In this section, we discuss the details of the state-of-the-art VLM models that have been used for the selected image classification and object detection tasks


\subsection{Image classification task}
In this study, the first image classification task involves congestion detection i.e., detecting the congestion in any of the highway lanes and classifying the image as congested or not. The second task of crack detection implies that, given the pavement surface images, the model needs to identify whether any cracks are present or not.

For these image classification tasks, the three models used are OpenAI Contrastive Language-Image Pre-Training (CLIP) \cite{clip}, Bootstrapping Language-Image Pre-training (BLIP) \cite{li2022blip}, and Large Language and Vision Assistant - Next Generation (LLaVA-NeXT)  \cite{llava_next} and GPT-4o. As these models have strong zero-shot performance, this eliminates the need for annotated training data in the image classification task. Therefore these models were selected for classifying vision-based transportation-related tasks using zero-shot prompting. As explained earlier, zero-shot prompting is a technique where a model is given a task or instruction without any prior example or training on that specific task.

\subsubsection{CLIP model}

CLIP model \cite{clip} is trained on 400 million image-text pairs available on the internet, allowing it to learn a range of visual features along with their corresponding text description. During the training of CLIP, it employs contrastive learning where it learns to predict which text and image are paired together. An image and text encoder are trained to maximize the cosine similarity of the correct image and text pairs while minimizing the cosine similarity of the incorrect pairings.
The CLIP model has zero-shot learning capability, allowing it to classify images based on natural language prompts without requiring additional task-specific training. 

In this study, image classification has been performed for two vision-based transportation tasks, congestion and crack classification. For both tasks, we used the names of the binary classes as the probable text pairings and employed CLIP to predict the most likely (image, text) pair.

To classify congestion, we used five different class names for the CLIP model: 

A1: ["Congested", "Non-congested"], 

A2: ["Congested lanes", "Non-congested lanes"], 

A3: ["Lanes with congestion", "Lanes without congestion"], 

A4: ["Queued traffic", "Free-flow traffic"], 

A5: ["Congested lanes", "Free-lanes"]. 

Similarly, for classifying cracks in the pavement surface, the different class names used were: 

B1: [“Cracked”, “Non-Cracked”], 

B2: [“Cracks present”, “Cracks absent”], 

B3: [“Cracked surface”, “Non-Cracked surface”], 

B4: [“Cracked pavement”, “Crack-free pavement”], 

B5: [“Crack”, “No crack”].

\subsubsection{BLIP model}
BLIP addresses both vision-language understanding and generation tasks by using a multimodal mixture of encoder-decoder architecture and a novel data bootstrapping technique called CapFilt. BLIP generates synthetic captions and filters out noisy ones to enhance training data quality, leading to state-of-the-art performance across various vision-language tasks, including image-text retrieval, image captioning, and visual question answering \cite{li2022blip}. Therefore BLIP is more versatile for both generation-based and understanding tasks compared to CLIP which focuses on alignment and representation learning.

Similar to the CLIP model, probable class names were used by BLIP to classify images for both tasks. The different class names utilized in the BLIP model are the same as those mentioned for the CLIP model.

\subsubsection{LLaVA model}

The Large Language and Visual Assistant (LLaVA)\cite{liu2024llava} is an open-source multimodal model that is designed to interpret and generate results based on both visual and textual input (10). It leverages the LLaMa \cite{llama} model and incorporates the pre-trained CLIP visual encoder for processing visual content. The encoder extracts visual features from input images and links them to language embedding through a trainable projection matrix, effectively translating visual features into language embedding tokens and bridging the gap between text and images. Although trained on smaller datasets than closed-source multimodal GPT models, Llava purports to demonstrate behavior analogous to the proprietary models.  

The LLaVA-NeXT (an updated version of LLaVa) \cite{llava_next} focuses on enhancing multimodal instruction following capabilities on data generated to follow detailed visual and textual instructions, for interactive and complex visual tasks. In contrast, CLIP learns generalizable visual representations from large-scale natural language supervision aligning image and text embedding to enable zero-shot learning across diverse vision. 

The LLaVA-NeXT model was also employed in this study for both classification tasks with different task-specific instructions. We first selected the initial prompt to query the model to generate the description of each image. To get the desired output as a discrete class name, we further queried the model with the generated description and follow-up prompt to get output as class names.

The five different initial prompts that were adopted to generate the description of congested/non-congested dataset are as follows:

P1: Classify whether highway lanes are congested or not in the image.

P2: Classify whether highway lanes are congested or not in the image.

P3: Classify whether in the image highway lanes are congested or not.

P4: Classify whether the highway have congested lane or free-lane in the image.

P5: Check whether the highway lanes are congested or not in the image.
\newline

The follow-up prompt selected corresponding to each of these initial prompts were

F1: Write Yes for congested, No for non-congested. 

F2: Write Congested lanes if lanes are congested, Free-lanes if lanes are not congested. 

F3: Write Congested lanes if lanes are congested, Free-lanes if lanes are not congested.

F4: Write  Congested lanes if lanes are congested, Free-lanes if free-lane. 

F5: Write Congested lanes if lanes are congested, Free-lanes if lanes are not congested. 
\newline

\noindent Similarly, for classifying cracked/non-cracked images, the initial prompt adopted were

P1: Classify whether the pavements have cracks or not in the image? 

P2: Classify whether the cracks are present or not in the pavement surface image?

P3: Classify whether the pavement surface is cracked or not in the image?

P4: Classify whether in the image, the pavement surface have cracks or not? 

P5: Check whether the pavement surface has any cracks or not?
\newline

The corresponding follow-up prompt selected for query were

F1: Write Cracked if cracks present, Non-cracked if cracks not present. 

F2: Write Cracked if cracks present, Non-cracked if cracks not present. 

F3: Write Cracked if surface is cracked, Non-cracked if surface is not-cracked.

F4: Write Cracked if surface has cracks, Non-cracked if surface do not have cracks.

F5: Write Cracked if cracks present, Non-cracked if cracks not present.

\subsubsection{GPT-4o}
GPT-4-o, with the "o" for "omni", is an advanced iteration of the Generative Pre-trained Transformer series by OpenAI. GPT-4o is built to handle multimodal data including text, images, and audio, allowing it to process and generate not only natural language but also interpret and respond to visual and auditory data. Its ability to understand and generate human-like text makes it a valuable tool in diverse fields.

In this study, we have prompted GPT-4o for both image classification tasks for comparison and evaluation purposes. The prompt used for congestion classification was \textit{"Can you tell me whether the closer lane are free lanes or not. Only return non-Congested if there are all free lanes otherwise return congested"}, whereas for crack classification \textit {"Can you tell me whether the pavements have cracks or not in the image. Only return yes if crack is present and no if crack is not present."}

\subsubsection{Benchmark CNN model}
The task of congestion classification has been compared using a DCNN model present in Chakraborty et al.\cite{chakraborty2018traffic}. It took 25 minutes to train the model on an NVIDIA Tesla K20m GPU with 4 GB RAM. On the other hand, for crack classification, the VLM models were compared with CNN EfficientNet B1 model architecture, it took 903 secs to run on the test dataset.

\subsection{Object detection tasks}

Our study focuses on the object detection task of identifying helmet violations. The task aims to detect whether motorcyclists were wearing helmets, which is a mandatory rule of road safety in many countries. This class is not present in the COCO dataset or any other pre-trained CNN model, necessitating fine-tuning of the model for our specific use case. Our objective is to explore how well zero-shot Vision Language Models (VLM) perform in these scenarios, to reduce the need for intensive datasets and fine-tune them, thereby streamlining the resource-intensive processes of dataset creation. 

We aim to detect two classes: “Helmet” – a motorcyclist wearing helmet, and “No-Helmet” – a motorcyclist without wearing helmet. While identifying the positive sentiment class is relatively straightforward, but the challenge lies in identifying the negative sentiment class, "No-Helmet," especially for Vision Language models. For any language model, it is easy to understand the positive sentiment class, but VLMs have been found to face difficulties identifying a negative sentiment class.  

For zero-shot object detection, we have been used the Vision Transformers for Open-World Localization (OWL-ViT) model \cite{OWL_vit} with basic classes and performing required post-processing to improve results. Additionally, we are utilizing textual class prompts to eliminate the need for post-processing. However, OWL-ViT does not perform well on textual classes since it is not trained on Large Language Models. As a result, we are considering open-source Large Language Vision models such as Llava-Next \cite{llava_next}, as well as close-source VLMs like GPT-4o \cite{gpt4o}.

\subsubsection{OWL-ViT model}
OWL-ViT \cite{OWL_vit} is a state-of-the-art open vocabulary object detection model, which was launched by the Google research team in 2022. This model is designed to understand the relationship between images and text. Operating as a zero-shot object detection model, it leverages CLIP \cite{clip} as its multimodal backbone in conjunction with a ViT-like (Vision Transformer-like) model. To use CLIP for object detection, OWL-ViT removes the final token pooling layer of the vision model and adds a lightweight classification and box head to each transformer output token pool. Open-vocabulary classification is achieved by substituting the fixed classification layer weights with the embeddings obtained from the text model. 

In this part of our study, we are focused on Helmet violation detection using OWL-ViT \cite{OWL_vit}. For that, we have explored its performance in basic classes, i.e., one-word classes given to the prompts. This method requires post processing for better accuracy. Furthermore, we have extended it to beyond one-word classes by checking its performance on textual classes.\\ 
\par

\noindent a) Detection of basic classes and post processing 

Initially, a single word class name is input via prompt, i.e., Motorbike, Person and Helmet. After detecting these classes, post processing is necessary to obtain the desired output class, i.e., (1) “Helmet” (a person who is sitting on a motorbike wearing helmet) and (2) “NoHelmet” (a person who is sitting on a motorbike without wearing helmet). 

Three steps are involved in the post-processing module, as shown in Fig 1. First, a non-maximum suppression method is used to remove the duplicated bounding boxes. Second, the person bounding boxes that are aligned with the motorbike are selected based on the calculation of the Intersection over Union (IoU) of the motorbike and person bounding boxes. Those with an IoU greater than 60\% are retained, thereby excluding persons not seated on motorbikes such as pedestrians, bicyclists etc. In the third step, we identify person bounding boxes sharing an IoU of over 60\% with the helmet bounding boxes and assign them to the "Helmet" class. Any remaining person bounding boxes, which are not aligned with the helmet bounding boxes, are categorized as members of the "NoHelmet" class. 

\begin{figure}[h]
\centering
\includegraphics[width=\textwidth]{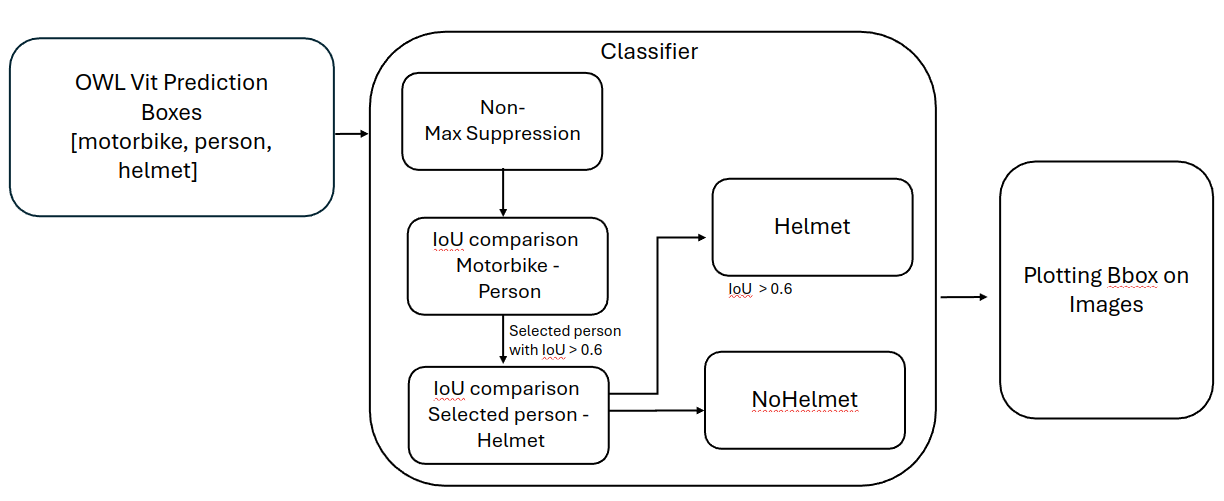}
\caption{OWL-ViT (basic prompt) with post-processing methodology}
\end{figure}

b) Detection using text classes directly so that post processing not required

As we had observed in the section above, OWL-ViT \cite{OWL_vit} with basic classes need some post processing. We aim to use VLMs so that no post-processing is needed. In this case, we provided prompt with textual classes, which consists of entire sentences instead of individual words, providing a complete explanation. OWL-ViT is an open vocabulary object detector, and it performs well at identifying basic classes. In our study we are interested in determining whether OWL-ViT could perform equally well with textual classes without requiring any post-processing. 

For our research we had selected different textual prompts 

Prompt 1: “A person on a motorbike wearing helmet” 

Prompt 2: “A person on a motorbike bareheaded” 

Prompt 3: “A person on a motorbike without wearing helmet” 

\subsubsection{LLava model}
As discussed in the earlier section, OWL-ViT \cite{OWL_vit} needs a post processing module to understand negative sentiment and give better results. In this scope of study, we wanted to test the performance of large language models like Llava-Next \cite{llava_next} on our use case. As mentioned in the section above, Llava-Next holds very good image understanding capabilities.  

In this scope of our study, we undertook several experiments with Llava-Next \cite{llava_next} to evaluate its performance in the field of Object Detection. To assess its performance in Image Understanding, we provided the prompt “\textit{Describe the image}”. The Llava-Next model is known for its exceptional image interpretation capabilities, yielding precise results. However, Llava-Next is unable to generate bounding boxes or provide updated images with bounding boxes. We attempted to obtain the coordinates using different prompts but found that it exhibits poor object localization capability and negative sentiment understanding. Therefore, to leverage Llava-Next image understanding capabilities, we combined OWL-ViT with Llava-Next. 

OWL-ViT (basic classes) with Llava-Next :
In this part of our study, we integrated OWL-ViT \cite{OWL_vit} with Llava-Next \cite{llava_next} to optimize the outcomes. Initially OWL-ViT was employed with basic classes in it’s prompt, specifically, Motorbike, Person and Helmet. Following this, we implemented non-maximum suppression to remove redundant bounding boxes. Subsequently, we utilized IoU selection to extract the images of individuals seated on motorbikes, as mentioned in the preceding section. These images were then processed as inputs for Llava-Next, using the prompt: \textit{"Identify whether all person sitting on motorbike is wearing helmet or not?"}. 

Additionally, we employed a follow-up prompt to assign discrete classes. Specifically, we assigned the class “Helmet” to all crop images in which all visible individuals were wearing helmets, and the class “NoHelmet” to the crop images in which any of the visible persons was not wearing a helmet. The follow-up prompt was: \textit{" Write no if any person is not wearing helmet and write yes if all person is wearing helmet."}  

\subsubsection{GPT-4o}
As mentioned above, GPT-4o \cite{gpt4o} is OpenAI latest LLM model. Being a closed source model, it exhibits visual language understanding than any other available model. It is exceptionally good in visual understanding, but similar to Llava-Next \cite{llava_next}, GPT-4o \cite{gpt4o} also lacks the capability of object localization and fails to return correct bounding box coordinates. Therefore, we combine OWL-ViT and GPT-4o. The crops from OWL-ViT (basic classes) are given to GPT-4o, with a prompt: “\textit{Can you tell me the if there is a person wearing helmet or not. Only return helmet if all person are wearing helmet otherwise result nohelmet}”. We don’t require a follow-up while using GPT-4o, as its textual understanding is good and it returned the expected classes, i.e., Helmet and NoHelmet.

\subsubsection{Benchmark CNN Model}
For comparing the results of VLM with CNN models, we finetune a YOLOv8 model \cite{yolov8}. The model is trained using 2500 training images. The YOLOv8 is trained on Nvidia-RTX A4000 GPU, it takes around 6 hours for 250 epochs.   

\section{Dataset}
In this section, we discussed the dataset used in our study for the tasks discussed earlier, starting with image classification tasks i.e., congestion and crack and classification and then object detection.

\subsection{Classification tasks}
The congestion dataset was taken from the work of Chakraborty et al \cite{chakraborty2018traffic}, where images were obtained from 121 cameras from the Iowa DOT CCTV camera database spread across the inter-states and highways. The dataset has 1010 images in total, having 516 congested images and 494 non-congested images of highways. 

The dataset used for crack classification is SDNET2018 \cite{dorafshan2018sdnet2018}, a publicly available dataset containing more than 56,000
images of concrete walls, bridges, and pavements. The pavement images are labeled and categorized into cracked and non-cracked classes. In our study, all the cracked 2608 images were used and we randomly selected 2600 non-cracked pavement images to balance the dataset.

\subsection{Object detection tasks}
The dataset used is sourced from AICity Challenge 2024, specifically Track 5 - Detecting Violation of Helmet Rule for Motorcyclists \cite{wangcityllava}. The dataset consists of 100 training and 100 testing videos, recorded at 10 fps and 1080p resolution from various locations in an Indian city. We extracted the dataset and selected 2500 training images and 200 test images. These images show a close-up view of traffic captured by cameras. The dataset contains 3 object classes: motorbike, Helmet (a person seated on a motorbike wearing a helmet), and NoHelmet (a person seated on a motorbike without wearing a helmet). The dataset includes images of individuals wearing helmets, scarves, and turbans, as well as those without any headgear. Additionally, it contains footage from congested lanes. The same test dataset, consisting of 200 images, has been utilized for the vision-language model OWL-ViT \cite{OWL_vit}, Llava-Next \cite{llava_next} and GPT-4o \cite{gpt4o}.

\section{Results and Discussion}
In this section, we discussed the results achieved in our study by applying different state-of-the-art VLM models discussed earlier, starting with image classification tasks i.e., congestion and crack and classification and then object detection.
\subsection{Classification tasks}
\subsubsection{Congestion classification}

\begin{figure}[!ht]
    \centering
    \begin{minipage}{0.9\textwidth}
        \centering
        \includegraphics[width=\textwidth]{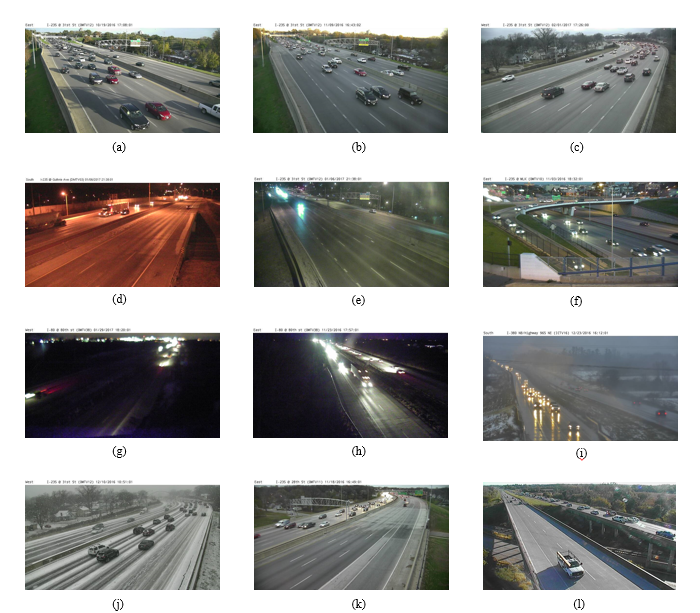}
        \caption{Result of congestion classification (a–c) true positives, (d–f) true negatives, (g–h)false positives,  (j–l)false negatives from CLIP, BLIP, and LlaVA-NeXT, respectively.}
    \end{minipage}
\end{figure}

The task of classifying congestion was accomplished by applying zero-shot prompting to the three models. As zero-shot prompting involves performing tasks without any training on those tasks, it has to rely entirely on the instruction provided in the prompt itself to perform the task. 

The performance of the CLIP model varied depending on the class name used as prompts. The accuracy achieved for different classnames A1, A2, A3, A4 and A5 were 76\%, 76\%, 66\%, 77\%, and 88\% respectively. The same classname was used by the BLIP model too but it had different results. For BLIP model the accuracy received were 86\%, 94\%, 93\%, 49\%, and 87\% for A1, A2, A3, A4, and A5 respectively. The LLaVA-NeXT model utilized the initial and follow-up prompts instead of class names to classify the image. The result achieved by different combinations of initial and follow-up prompts, P1-F1, P2-F2, P3-F3, P4-F4, P5-F5 was 86\%, 87\%, 82\%, 64\%, and 87\% accuracy respectively. The best results of all the models are presented in Table 1 with the Precision, Recall, and F1-score. 

\begin{table}[htbp]
  \centering
  \caption{Best performance metrics for different models for congestion classification task}
    \begin{tabular}{|l|l|l|l|l|l|}
    \hline
    \textbf{Model} & \textbf{Prompt used} & \textbf{Precision} & \textbf{Recall} & \textbf{F1-score} & \textbf{Inference time per image} \\\hline
    CLIP  & A5 & 0.85  & 0.93  & 0.89  & 0.43 sec \\\hline
    BLIP  & A2 & 0.95  & 0.92  & 0.94  & 0.49 sec \\\hline
    LLaVA-NeXT & P2-F2 & 1     & 0.81  & 0.87  & 5.4 sec \\\hline
    DCNN \cite{{chakraborty2018traffic}} &       & 0.87  & 0.94  & 0.9   & 0.05 sec \\\hline
    GPT-4o &       & 0.88  & 0.84  & 0.86  & 1.5 sec \\\hline
    \end{tabular}%
  \label{tab:performance_metrics}%
\end{table}%


            
                     


                       


 The True Positive (TP), True Negative (TN), False Positive (FP), and  False Negative (FN) are shown in Figure 3. As demonstrated in Fig 2 (g-i), it was observed that the models gave false positive results for the night-time because of the lightning effect and hence assumed it to be congested. Moreover, it can be also inferred from Fig 2 (j-l) that if one of the lanes had free-flowing conditions, it was classified as non-congestion even though the other lanes had congestion. The models are still not capable of understanding language in terms of whether any of the lanes are congested or not.  
From Table 1, it can be inferred that all the models gave comparative results and BLIP even outperformed the benchmark DCNN model.


\subsubsection{Crack classification}
\begin{figure}[!ht]
    \centering
    \begin{minipage}{0.8\textwidth}
        \centering
        \includegraphics[width=\textwidth]{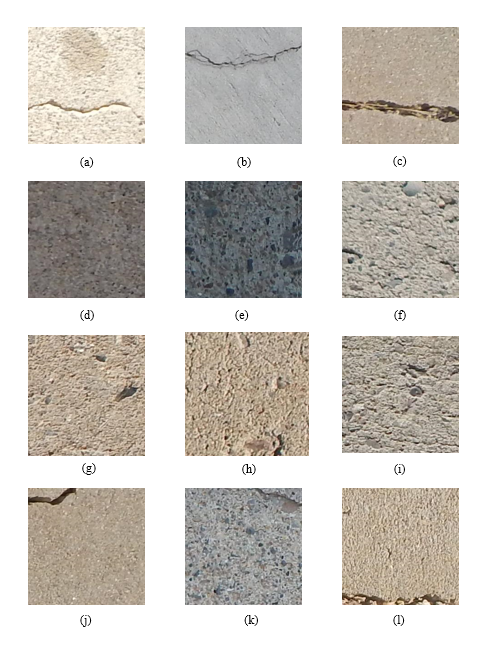}

        \caption{Result of crack classification using three models (a–c) true positives, (d–f) true negatives, (g–h) false positives,  (j–l) false negatives from CLIP, BLIP, and LlaVA-NeXT, respectively.}
     \end{minipage}
\end{figure}

The zero-shot prompting techniques being highly flexible allows the model to be applied to a wide range of tasks. Therefore, we also applied it to the crack classification task, but its ability to adapt to the specific demands of a new task is limited by what it has learned during its training.

The prompts play a vital role in getting the result from models through zero-shot prompting. The different class names used in the CLIP model with their accuracy, B1, B2, B3, B4, and B5 with 79\%, 68\%, 79\%, 74\%, and 70\% respectively. The same class name was used by the BLIP model and it has varying results to CLIP, the accuracy achieved was 50\%, 71\%, 57\%, 61\%, and  50\%, for class names B1, B2, B3, B4, and B5 respectively. Though the class names used by both models were the same, the results attained were contrasting. For the first and the last prompt used in the BLIP, it gave 50\% accuracy which represents that the model is not able to handle negation prompts. The initial and follow-up prompts were used by the LLaVA-NeXT model to classify the image. The result attained by different combinations of initial and follow-up prompts P1-F1, P2-F2, P3-F3, P4-F4, and P5-F5 was 72\%, 67\%, 53\%, 58\%, 76\% accuracy respectively. 

\begin{table}[htbp]
  \centering
  \caption{Best performance metrics for different models for crack classification task}
    \begin{tabular}{|l|l|l|l|l|l|}
    \hline
    \textbf{Model} & \textbf{Prompt used} & \textbf{Precision} & \textbf{Recall} & \textbf{F1-score} & \textbf{Inference time per image} \\
    \hline
    CLIP  & B1    & 0.81  & 0.77  & 0.78  & 0.44 sec \\
    \hline
    BLIP  & B2    & 0.65  & 0.95  & 0.77  & 0.47 sec \\
    \hline
    LLaVA-NeXT & P5-F5 & 0.86  & 0.81  & 0.84  & 5.2 sec \\
    \hline
    CNN   &      & -     & -     & 0.86  & 0.06 sec \\
    \hline
    GPT-4o &      & 0.56  & 0.81  & 0.67  & 1.5 sec \\
    \hline
    \end{tabular}%
  \label{tab:performance_metrics_crack}%
\end{table}%


The best result of all the models is demonstrated with precision, recall, and F1-score of each class in Table 2. The TP, TN, FP, and  FN are shown in Figure 3. It was inferred that the models are not able to distinguish between rough surface and crack, all the models classified the rough surface as cracked as shown in Fig 3 (g-h). On the other hand, if the crack was present near the edge as shown in Fig 3 (j-l), the model was not able to identify the crack.

In the image classification task, VLMs performed well even with zero-shot prompting. As no specific prompt was used, further by applying prompt-based strategies, we can enhance the models’ performance, making them viable for high-level tasks and possibly reducing the reliance on extensively annotated datasets.


\subsection{Object detection tasks}

\subsubsection{OWL-ViT}

\begin{figure}[!ht]
    \centering
    \begin{minipage}{0.9\textwidth}
        \centering
        \includegraphics[width=\textwidth]{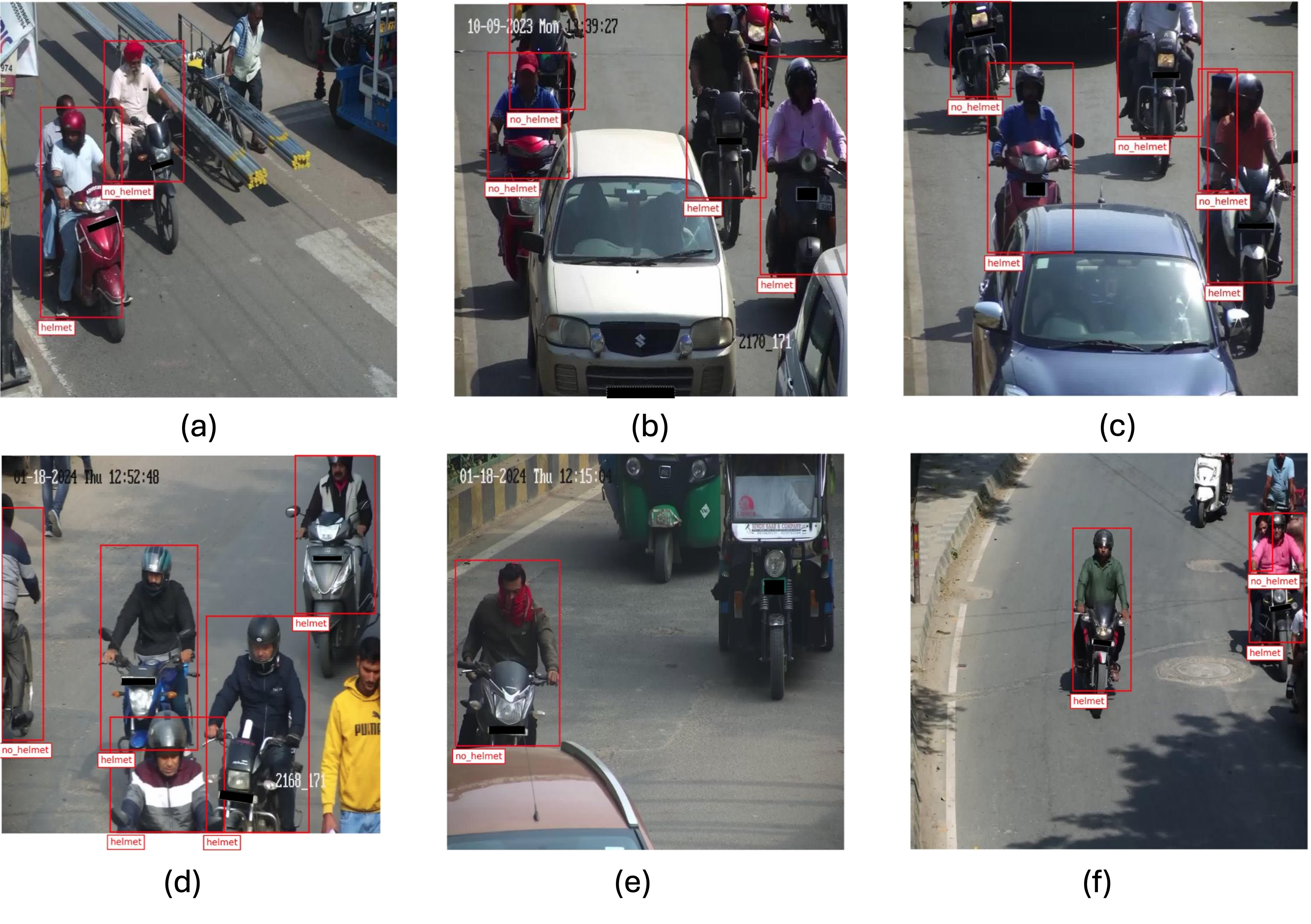}


    \end{minipage}

    \caption{Result of Helmet Violation Detection, using OWL-ViT with Post-processing }
    \label{fig:subfigures_post}
\end{figure}

As mentioned in the Table 3, OWL-ViT \cite{OWL_vit} shows significant good results when given basic classes in its prompts, such as Motorbike, Person, and Helmet. In order to achieve the Helmet and NoHelmet classes, some post-processing needs to be done. After post-processing, the Zero-Shot OWL-ViT model shows significantly better results compared to a trained YOLOv8 \cite{yolov8} model. It achieved a precision of 95\% in the Helmet class and 74\% precision in the NoHelmet class, as mention in Table 3. The inference results are shown in the Fig 4. One major advantage of the OWL-ViT is, with basic prompts, can also identify the differences between a cap, turban, scarf, and helmet, as shown in the Fig 4 (a,b,e).

OWL-ViT \cite{OWL_vit} has not been trained on higher language models, resulting in lack of visual and textual understanding. With the help of different prompts, we have observed that OWL-ViT performs poorly in processing textual data that consist of complete sentences. Furthermore, it also lacks the understanding of negative sentiments.

\subsubsection{LlaVa-Next}
Llava-Next \cite{llava_next} utilizes LlaMa models \cite{llama}, showing significant advancements in image understanding. Llava-Next cannot detect objects or provide images with bounding boxes, which can be achievable by OWL-ViT \cite{OWL_vit}. In our study, by tuning prompt, we get the bounding box coordinates in pascal voc format \cite{pascal} (xmin, ymin, xmax, ymax). Llava-Next cannot accurately locate objects and gives incorrect coordinates, despite its excellent understanding of prompts and images. 

Our approach involved integrating the OWL-ViT \cite{OWL_vit} model with Llava-Next \cite{llava_next} to leverage its image understanding capabilities. The basic class prompts of OWL-ViT were utilized as inputs for the Llava-Next model, as shown in Fig 5(a). Furthermore, we employed follow-up prompt for summarizing the results. As mentioned in the Table3, with our experiment we achieved 88\% precision in Helmet class and 90\% precision in NoHelmet class. 

\subsubsection{GPT-4o}
GPT-4o \cite{gpt4o} has a great capability for understanding visual language, similar to Llava-Next \cite{llava_next}. However, it does not support object detection. It has excellent image understanding, even more so than Llava-Next. By combining it with OWL-ViT, and providing GPT-4o with crops of OWL-ViT \cite{OWL_vit} (basic class prompt), we achieved 99\% precision in the Helmet class and 92\% precision in the NoHelmet class, which is almost equal to the benchmark results of CNN, which is shown in the Fig 5(b). Additionally, GPT-4o shows a recall of 99\% for the NoHelmet class, which is higher than the benchmark results of CNN as shown in Table 3.  \\
\par

\begin{figure}
    \centering
    \begin{minipage}{0.8\textwidth}
        \centering
        \includegraphics[width=\textwidth]{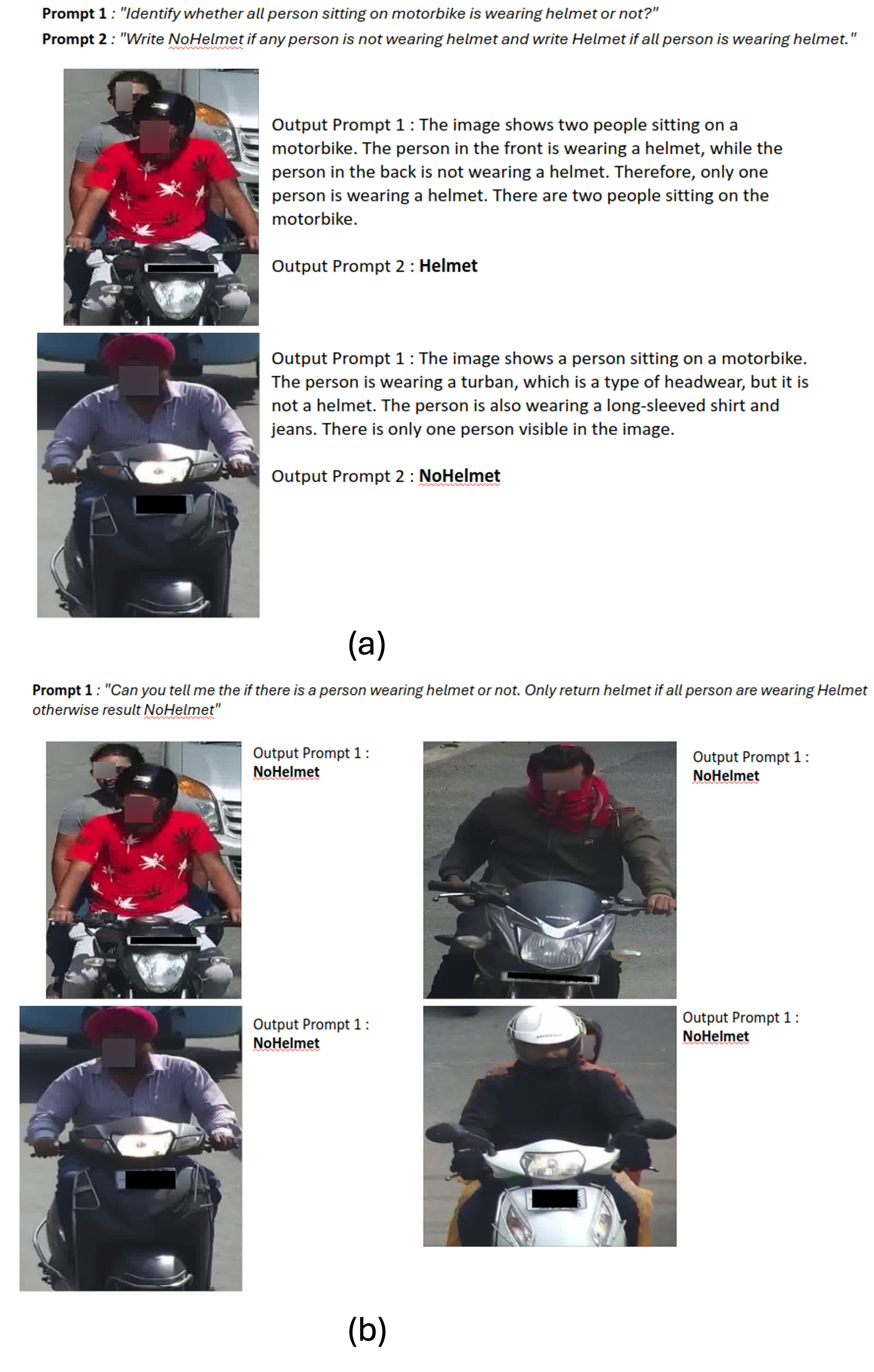}
    \end{minipage}
    \hfill

    \caption{Result of Helmet Violation Detection, (a) using OWL-ViT with Llava-Next (b) using OWL-ViT with GPT-4o }
    \label{fig:subfigures}
\end{figure}

\noindent According to table 3, VLM demonstrates significantly good precision and recall in zero-shot performance. This capability can lead to cost reduction by eliminating the need to fine-tune models for every new use case and annotate millions of images. With the right methodology, engineering, and utilization, VLM has the potential to outperform traditional CNN models. Despite excelling in these areas, VLM has poor image localization capabilities. Models such as Llava-Next and GPT-4o understand images well but struggle to localize objects. Additionally, VLMs are not currently lightweight or fast compared to CNN models. For example, OWL-ViT with Llava-Next took 6.2 seconds to process one image, while OWL-ViT with post-processing took 0.68 seconds and even the closely related GPT-4o took 3.3 seconds. In contrast, a CNN model only took 0.14 seconds for the same task. While VLMs can provide better accuracy, further work is needed to make them suitable for real-time field use.\\

\begin{table}[htbp]
    \centering
        \caption{Best performance metrics for different models for object detection task results}
    \begin{tabular}{|c|c|c|c|c|l|} \hline  
         \textbf{S.No.}&  \textbf{Model Name}&  \textbf{Precision}&  \textbf{Recall}& \textbf{F1-Score}  &\textbf{Inference time per image}\\ \hline  
         \multicolumn{5}{|l|}{Class - \textbf{Helmet}} &\\ \hline  
         1.&  YoloV8&  0.99&  0.98& 0.98 &0.14 sec\\ \hline  
         2.&  OWL-ViT with post processing&  0.96&  0.94& 0.95 &0.68 sec\\ \hline  
         3.&  OWL-ViT with Llava-Next&  0.88&  0.98& 0.93 &6.2 sec\\ \hline  
         4.&  OWL-ViT with GPT-4o&  0.99&  0.96&  0.97&3.3 sec\\ \hline  
         \multicolumn{5}{|c|}{} &\\ \hline  
         \multicolumn{5}{|l|}{Class - \textbf{NoHelmet}} &\\ \hline  
         1.&  YoloV8&  0.96&  0.90& 0.93 &0.14 sec\\ \hline  
 2.& OWL-ViT with post processing& 0.74& 0.80&0.77 &0.68 sec\\ \hline  
 3.& OWL-ViT with Llava-Next& 0.90& 0.55&0.68 &6.2 sec\\ \hline  
 4.& OWL-ViT with GPT-4o& 0.92& 0.99& 0.96&3.3 sec\\ \hline 
    \end{tabular}

    \label{tab:my_label}
\end{table}

\section{Conclusions}
The objective of the study was to understand the performance of visual language models in vision-based transportation tasks. It was carried out by comparing different state-of-the-art VLM models by zero-shot prompting for two tasks i.e., image classification and object detection. The transportation-related vision task selected for image classification was congestion detection and crack identification whereas, for object detection, it was the identification of helmet violations. VLMs performed at par sometimes but performance needs to be improved in terms of prompt engineering, localization of object detected, and inference time. This paper gives a comprehensive understanding of VLM models' limitations which need to be worked upon for large-scale implementation. Future studies can also be done in other case studies related to vision-based transportation tasks. Further other VLM models can also be explored with different benchmark datasets.\\

 \section{Author Contributions}
The authors confirm their contribution to the paper as follows: study conception and design:  T. Singh, S. Prajapati, C. Hedge, P. Chakraborty; data collection: S. Prajapati, T. Singh; analysis and interpretation of results: S. Prajapati, T. Singh, C. Hedge, P. Chakraborty; draft manuscript preparation: S. Prajapati, T. Singh, C. Hedge, P. Chakraborty. All authors reviewed the results and approved the final version of the manuscript.\\

\section{Acknowledgements}
Our research results are based upon work supported by the IITK-NYU Joint Research Grant. Any opinions, findings, and conclusions, or recommendations expressed in this material are those of the author(s) and do not necessarily reflect the views of the IITK and NYU. 

\bibliographystyle{trb}
\bibliography{trb_template}

\begin{thebibliography}{20}
\providecommand{\natexlab}[1]{#1}

\bibitem[{Bordes et~al.(2024)Bordes, Pang, Ajay, Li, Bardes, Petryk, Ma{\~n}as, Lin, Mahmoud, Jayaraman et~al.}]{vlm}
Bordes, F., R.~Y. Pang, A.~Ajay, A.~C. Li, A.~Bardes, S.~Petryk, O.~Ma{\~n}as, Z.~Lin, A.~Mahmoud, B.~Jayaraman, et~al., An introduction to vision-language modeling. \emph{arXiv preprint arXiv:2405.17247}, 2024.

\bibitem[{Lin et~al.(2014)Lin, Maire, Belongie, Hays, Perona, Ramanan, Doll{\'a}r, and Zitnick}]{coco}
Lin, T.-Y., M.~Maire, S.~Belongie, J.~Hays, P.~Perona, D.~Ramanan, P.~Doll{\'a}r, and C.~L. Zitnick, Microsoft coco: Common objects in context. In \emph{Computer Vision--ECCV 2014: 13th European Conference, Zurich, Switzerland, September 6-12, 2014, Proceedings, Part V 13}, Springer, 2014, pp. 740--755.

\bibitem[{Radford et~al.(2021{\natexlab{a}})Radford, Kim, Hallacy, Ramesh, Goh, Agarwal, Sastry, Askell, Mishkin, Clark et~al.}]{clip}
Radford, A., J.~W. Kim, C.~Hallacy, A.~Ramesh, G.~Goh, S.~Agarwal, G.~Sastry, A.~Askell, P.~Mishkin, J.~Clark, et~al., Learning transferable visual models from natural language supervision. In \emph{International conference on machine learning}, PMLR, 2021{\natexlab{a}}, pp. 8748--8763.

\bibitem[{Li et~al.(2022)Li, Li, Xiong, and Hoi}]{li2022blip}
Li, J., D.~Li, C.~Xiong, and S.~Hoi, Blip: Bootstrapping language-image pre-training for unified vision-language understanding and generation. In \emph{International conference on machine learning}, PMLR, 2022, pp. 12888--12900.

\bibitem[{Minderer et~al.(2022)Minderer, Gritsenko, Stone, Neumann, Weissenborn, Dosovitskiy, Mahendran, Arnab, Dehghani, Shen et~al.}]{OWL_vit}
Minderer, M., A.~Gritsenko, A.~Stone, M.~Neumann, D.~Weissenborn, A.~Dosovitskiy, A.~Mahendran, A.~Arnab, M.~Dehghani, Z.~Shen, et~al., Simple open-vocabulary object detection. In \emph{European Conference on Computer Vision}, Springer, 2022, pp. 728--755.

\bibitem[{Li et~al.(2024)Li, Zhang, Zhang, Zhang, Li, Li, Ma, and Li}]{llava_next}
Li, F., R.~Zhang, H.~Zhang, Y.~Zhang, B.~Li, W.~Li, Z.~Ma, and C.~Li, LLaVA-NeXT-Interleave: Tackling Multi-image, Video, and 3D in Large Multimodal Models. \emph{arXiv preprint arXiv:2407.07895}, 2024.

\bibitem[{Radford et~al.(2021{\natexlab{b}})Radford, Kim, Hallacy, Ramesh, Goh, Agarwal, Sastry, Askell, Mishkin, Clark et~al.}]{gpt4o}
Radford, A., J.~W. Kim, C.~Hallacy, A.~Ramesh, G.~Goh, S.~Agarwal, G.~Sastry, A.~Askell, P.~Mishkin, J.~Clark, et~al., Learning transferable visual models from natural language supervision. In \emph{International conference on machine learning}, PMLR, 2021{\natexlab{b}}, pp. 8748--8763.

\bibitem[{Pan et~al.(2024)Pan, Yaman, Nesti, Mallik, Allievi, Velipasalar, and Ren}]{pan2024vlp}
Pan, C., B.~Yaman, T.~Nesti, A.~Mallik, A.~G. Allievi, S.~Velipasalar, and L.~Ren, VLP: Vision Language Planning for Autonomous Driving. In \emph{Proceedings of the IEEE/CVF Conference on Computer Vision and Pattern Recognition}, 2024, pp. 14760--14769.

\bibitem[{Tian et~al.(2024)Tian, Gu, Li, Liu, Hu, Wang, Zhan, Jia, Lang, and Zhao}]{tian2024drivevlm}
Tian, X., J.~Gu, B.~Li, Y.~Liu, C.~Hu, Y.~Wang, K.~Zhan, P.~Jia, X.~Lang, and H.~Zhao, Drivevlm: The convergence of autonomous driving and large vision-language models. \emph{arXiv preprint arXiv:2402.12289}, 2024.

\bibitem[{Xu et~al.(2023)Xu, Zhang, Xie, Zhao, Guo, Wong, Li, and Zhao}]{xu2023drivegpt4}
Xu, Z., Y.~Zhang, E.~Xie, Z.~Zhao, Y.~Guo, K.~K. Wong, Z.~Li, and H.~Zhao, Drivegpt4: Interpretable end-to-end autonomous driving via large language model. \emph{arXiv preprint arXiv:2310.01412}, 2023.

\bibitem[{Hu et~al.(2023)Hu, Ou, Wang, and Yu}]{hu2023VLN}
Hu, Y., D.~Ou, X.~Wang, and R.~Yu, Enabling Vision-and-Language Navigation for Intelligent Connected Vehicles Using Large Pre-Trained Models. In \emph{2023 IEEE International Conferences on Internet of Things (iThings) and IEEE Green Computing \& Communications (GreenCom) and IEEE Cyber, Physical \& Social Computing (CPSCom) and IEEE Smart Data (SmartData) and IEEE Congress on Cybermatics (Cybermatics)}, IEEE, 2023, pp. 390--396.

\bibitem[{Wang et~al.(2024)Wang, Anastasiu, Tang, Chang, Yao, Zheng, Rahman, Arya, Sharma, Chakraborty et~al.}]{wangcityllava}
Wang, S., D.~C. Anastasiu, Z.~Tang, M.-C. Chang, Y.~Yao, L.~Zheng, M.~S. Rahman, M.~S. Arya, A.~Sharma, P.~Chakraborty, et~al., The 8th AI City Challenge. In \emph{Proceedings of the IEEE/CVF Conference on Computer Vision and Pattern Recognition}, 2024, pp. 7261--7272.

\bibitem[{Hasan et~al.(2024)Hasan, Chen, Wang, Rahman, Joshi, Velipasalar, Hegde, Sharma, and Sarkar}]{hasan2024driveclip}
Hasan, M.~Z., J.~Chen, J.~Wang, M.~S. Rahman, A.~Joshi, S.~Velipasalar, C.~Hegde, A.~Sharma, and S.~Sarkar, Vision-language models can identify distracted driver behavior from naturalistic videos. \emph{IEEE Transactions on Intelligent Transportation Systems}, 2024.

\bibitem[{Lv and Sun(2024)}]{lv2024VAD}
Lv, H. and Q.~Sun, Video anomaly detection and explanation via large language models. \emph{arXiv preprint arXiv:2401.05702}, 2024.

\bibitem[{Liu et~al.(2024)Liu, Li, Wu, and Lee}]{liu2024llava}
Liu, H., C.~Li, Q.~Wu, and Y.~J. Lee, Visual instruction tuning. \emph{Advances in neural information processing systems}, Vol.~36, 2024.

\bibitem[{Touvron et~al.(2023)Touvron, Lavril, Izacard, Martinet, Lachaux, Lacroix, Rozi{\`e}re, Goyal, Hambro, Azhar et~al.}]{llama}
Touvron, H., T.~Lavril, G.~Izacard, X.~Martinet, M.-A. Lachaux, T.~Lacroix, B.~Rozi{\`e}re, N.~Goyal, E.~Hambro, F.~Azhar, et~al., Llama: Open and efficient foundation language models. \emph{arXiv preprint arXiv:2302.13971}, 2023.

\bibitem[{Chakraborty et~al.(2018)Chakraborty, Adu-Gyamfi, Poddar, Ahsani, Sharma, and Sarkar}]{chakraborty2018traffic}
Chakraborty, P., Y.~O. Adu-Gyamfi, S.~Poddar, V.~Ahsani, A.~Sharma, and S.~Sarkar, Traffic congestion detection from camera images using deep convolution neural networks. \emph{Transportation Research Record}, Vol. 2672, No.~45, 2018, pp. 222--231.

\bibitem[{Terven et~al.(2023)Terven, C{\'o}rdova-Esparza, and Romero-Gonz{\'a}lez}]{yolov8}
Terven, J., D.-M. C{\'o}rdova-Esparza, and J.-A. Romero-Gonz{\'a}lez, A comprehensive review of yolo architectures in computer vision: From yolov1 to yolov8 and yolo-nas. \emph{Machine Learning and Knowledge Extraction}, Vol.~5, No.~4, 2023, pp. 1680--1716.

\bibitem[{Dorafshan et~al.(2018)Dorafshan, Thomas, and Maguire}]{dorafshan2018sdnet2018}
Dorafshan, S., R.~J. Thomas, and M.~Maguire, SDNET2018: An annotated image dataset for non-contact concrete crack detection using deep convolutional neural networks. \emph{Data in brief}, Vol.~21, 2018, pp. 1664--1668.

\bibitem[{Hoiem et~al.(2009)Hoiem, Divvala, and Hays}]{pascal}
Hoiem, D., S.~K. Divvala, and J.~H. Hays, Pascal VOC 2008 challenge. \emph{World Literature Today}, Vol.~24, No.~1, 2009, pp. 1--4.

\end{thebibliography}

\end{document}